\documentclass{article}
\usepackage[numbers]{natbib}
\setcitestyle{authoryear,open={(},close={)}}
\usepackage{arxiv}
\usepackage{latexsym}
\usepackage{subfig}
\usepackage{caption}
\usepackage{graphicx}
\usepackage{mathptmx}

\usepackage{comment}

\usepackage{amsmath}
\usepackage{amsfonts}
\usepackage{amssymb}
\usepackage{amsbsy}
\usepackage{amsthm}
\usepackage{multirow}
\usepackage[ruled,vlined]{algorithm2e}
\DeclareMathOperator{\E}{\mathbb{E}}
\usepackage[pdftex,colorlinks=true,urlcolor=blue,citecolor=black,anchorcolor=black,linkcolor=black]{hyperref}

\newtheoremstyle{wsc}
{3pt}
{3pt}
{}
{}
{\bf}
{}
{.5em}
{}

\theoremstyle{wsc}

\begin{document}

%
%

\title{GREEN SIMULATION ASSISTED REINFORCEMENT LEARNING WITH MODEL RISK FOR BIOMANUFACTURING LEARNING AND CONTROL}

\author{\textbf{Hua Zheng, Wei Xie}
\\[12pt]
 Department of Mechanical and Industrial Engineering\\
 Northeastern University\\
 Boston, MA 02115, USA\\
 \and
\textbf{M. Ben Feng}\\[12pt]
 Department of Statistics and Actuarial Science\\
University of Waterloo \\
Waterloo, Ontario, CANADA\\
 }

\maketitle

\section*{ABSTRACT}

Biopharmaceutical manufacturing faces critical challenges, including complexity, high variability, lengthy lead time, and limited historical data and knowledge of the underlying system stochastic process. To address these challenges, we propose a green simulation assisted model-based reinforcement learning to support process online learning and guide dynamic decision making. Basically, the process model risk is quantified by the posterior distribution. At any given policy, we predict the expected system response with prediction risk accounting for both inherent stochastic uncertainty and model risk. Then, we propose green simulation assisted reinforcement learning and derive the mixture proposal distribution of decision process and likelihood ratio based metamodel for the policy gradient, which can selectively reuse process trajectory outputs collected from previous experiments to increase the simulation data-efficiency, improve the policy gradient estimation accuracy, and speed up the search for the optimal policy.  Our numerical study indicates that the proposed approach demonstrates the promising performance.

\section{INTRODUCTION}
\label{sec:intro}

\textit{To address critical needs in biomanufacturing automation, in this paper, we introduce a green simulation assisted Bayesian reinforcement learning to support bioprocess online learning and guide dynamic decision making.}
 The biomanufacturing industry is growing rapidly and becoming one of the key drivers of personalized medicine. However, biopharmaceutical production faces critical {challenges}, including complexity, high variability, long lead time, and very limited process data. 
 Biotherapeutics are manufactured in living cells whose biological processes are complex and have highly variable outputs (e.g., product critical quality attributes (CQAs)) whose values are determined by many factors (e.g., raw materials, media, critical process parameters (CPPs)).
 As new biotherapeutics (e.g., cell and gene therapies) become more and more ``personalized," biomanufacturing requires more advanced manufacturing protocols. 
In addition, the analytical testing time required by biopharmaceuticals of complex molecular structure is lengthy, and the process observations are relatively limited.

Driven by these challenges, we consider the model-based reinforcement learning (MBRL) or Markov Decision Process (MDP) to fully leverage the existing bioprocess domain knowledge, utilize the limited process data, support online learning, and guide dynamic decision making. At each time step $t$, the system is in state $s_t$, and the decision maker takes the action $a_t$ by following a policy $a_t=\pi_t(a_t|s_t)$. 
At the next time step $(t+1)$, the system evolves to new state $s_{t+1}$ by following the state transition probabilistic model $P(s_{t+1}|s_t,a_t;\pmb\omega)$, 
and then we collect a reward $r_t(a_t,s_t)$.
\textit{Thus, the statistical properties and dynamic evolution of stochastic control process depend on decision policy $\pi_t$ and state transition model $P(s_{t+1}|s_t,a_t;\pmb\omega)$.}
In the biomanufacturing, the prior knowledge of state transition model is constructed based on the existing biological/physical/chemical mechanisms and dynamics. The unknown model parameters $\pmb\omega$ (e.g., cell growth, protein production, and substrate consumption rates in cell culture; nucleation rate and heat transfer coefficients in freeze drying) will be online learned and updated as the arrivals of new process data. \textit{The optimal policy depends on the current knowledge of process model parameters.}

In this paper, we propose a green simulation assisted Bayesian reinforcement learning (GS-RL) 
to guide dynamic decision making. Given any policy, we predict the expected system response with prediction risk accounting for both process inherent stochastic uncertainty and model estimation uncertainty, call \textit{model risk}. The model risk is quantified by the posterior distribution and it can efficiently leverage the existing bioprocess domain knowledge through the selection of prior and support the online learning. \textit{Thus, the proposed Bayesian reinforcement learning can provide the robust dynamic decision guidance}, which can be applicable for cases with various amount of process historical data. 
In addition, motivated by the studies on green simulation (i.e.,~\cite{FengGreenSim2017} and~\cite{Dong2018}), we propose the stochastic control process likelihood ratio-based metamodel to improve the policy gradient estimation, which can fully leverage the historical trajectories generated with various state transition models and policies.
\textit{Therefore, the proposed green simulation assisted Bayesian reinforcement learning can: (1) incorporate the existing process domain knowledge; (2) facilitate the interprertable online learning; (3) guide complex bioprocess dynamic decision making; and (4) provide the reliable, flexible, robust, and coherent decision guidance.}

For the model-free inforcement learning,~\cite{mnih2015humanlevel} introduce the experience replay (ER) to reuse the past experience, increase the data efficiency, and decrease the data correlation. It randomly samples and reuses the past trajectories. Built on ER,~\cite{Schaul2016PrioritizedER} further propose the prioritized experience replay (PER), which prioritizes the historical trajectories based on temporal-difference error.

The \textbf{main contribution} of our study is to propose a green simulation assisted Bayesian reinforcement learning (GS-RL). 
Even though both GS-RL and PER are motivated by ``experience replay" and reuse the historical data, there is the fundamental difference between GS-RL and PER. In our approach, the posterior distribution of state transition model can provide the risk- and science-based knowledge of underlying bioprocess dynamic mechanisms, and facilitate the online learning.
Then, the likelihood ratio of stochastic decision process is used to construct the metamodel of policy gradient in the complex decision process space, accounting for the selection and impact of both policy and state transition model. 
It allows us to reuse the trajectories from previous experiments, and the weight assigned to each trajectory depends on its importance 
measured by the spatial-temporal distance of decision processes.
In addition, a mixture process proposal distribution used in the likelihood ratio can improve the estimation accuracy and stability of policy gradient and speed up the search for the optimal policy. Since the model risk is automatically updated during the learning, our approach can dynamically adjust the importance weights on the previous trajectories, 
which makes GS-RL flexible, efficient, and automatically deal with non-stationary bioprocess.

The organization of the paper is as follow. In Section~\ref{sec: GS based MDP}, we provide the problem description. To facilitate the biomanufacturing process online learning and automation, we focus on the model-based reinforcement learning with the posterior distribution quantifying the model risk. 
Then, in Section~\ref{sec: green simulation assisted RL with model risk}, we propose the green simulation assisted policy gradient, which can fully leverage the process trajectories obtained from previous experiments 
and speed up the search for the optimal policy. After that, a biomanfuacturing example is used to study the performance of proposed approach and compare it with the state-of-art policy gradient approaches in Section~\ref{Sec: empirical study}. We conclude this paper in Section~\ref{Sec: conclusion}. 


\section{PROBLEM DESCRIPTION AND MODEL BASED REINFORCEMENT LEARNING}
\label{sec: GS based MDP}

To facilitate the biomanufacturing automation, we consider the reinforcement learning for finite horizon problem. In Section~\ref{subsec:mdpFormulation}, we suppose
the underlying model of production process is known and review the model-based reinforcement learning. 
Since the process model is typically unknown and estimated by very limited process data in the biomanufacturing, 
in Section~\ref{subsec:ModelRisk}, the posterior distribution is used to quantify the model risk and the posterior predictive distribution, accounting for stochastic uncertainty and model risk, is used to generate the trajectories characterizing the overall prediction risk.
Thus, in this paper, we focus on the model-based reinforcement learning with model risk so that we can efficiently leverage the existing process knowledge, support online learning, and guide process dynamic decision making.


\subsection{Model-Based Reinforcement Learning for Dynamic Decision Making}
\label{subsec:mdpFormulation}

We formalize model-based reinforcement learning or Markov decision process (MDP) over finite horizon $H$ as $(\mathcal{S},\mathcal{A},P, r, s_1, H)$, where $\mathcal{S}$ is a set of states, $s_1$ is the starting state, $\mathcal{A}$ is the set of actions. The process proceeds in discrete time step $t = 1, 2, . . . ,H$. In each $t$-th time step, the agent observes the current state $s_t \in \mathcal{S}$, takes an action $a_t \in \mathcal{A}$,
and observes a feedback in form of a reward signal $r_{t+1} \in \mathbb{R}$. Moreover, let $\pi_{\pmb\theta}: \mathcal{S} \rightarrow \mathcal{A}$ denote a policy specified by parameter vector $\pmb\theta\in\mathbb{R}^d$. The policy is a function of current state, $a_t=\pi_{\pmb\theta}(s_t)$, whose output is action for deterministic policy or its selection
probabilities for random policy. 
For \textit{non-stationary} finite horizon MDP, we can write $\pmb{\pi}_{\pmb{\theta}}=(\pi_{\pmb{\theta}}^1,\ldots, \pi_{\pmb{\theta}}^H)$. 

Let $P(s_{t+1}|s_t,a_t;\pmb\omega^c)$ represent the state transition model characterizing the probability
of transitioning to a particular state $s_{t+1}$ from state $s_t$. Suppose the underlying process model can be characterized by parameters $\pmb\omega^c$. 
Let $D_{P_{\pmb\omega^c}}^{\pmb{\pi}_{\pmb{\theta}}}
(\pmb{\tau})$ denote the probability distribution of the trajectory

$$\pmb{\tau}= \pmb{\tau}_{[1:H-1]}\equiv (s_1,a_1,s_2,a_2,\ldots, s_{H-1},a_{H-1},s_H)$$ of state-action sequence over transition probabilities parameterized by transition model $P(s_{t+1}|s_t,a_t;\pmb\omega^c)$ starting from state $s_1$ and following policy $\pmb{\pi}_{\pmb{\theta}}$. 
\textit{The bioprocess trajectory length $H$ can be scenario-dependent.} For example, it can depend on the CQAs of raw materials and working cells.
We write the distribution of decision process trajectory as

\begin{equation}
 D_{P_{\pmb\omega^c}}^{\pmb{\pi}_{\pmb{\theta}}}(\pmb{\tau}) \equiv p(s_1;\pmb\omega^c)\prod^{H-1}_{t=1}\pi_{\pmb{\theta}}^t(a_t|s_t)p(s_{t+1}|s_t,a_t;\pmb\omega^c).
\end{equation}


Let $R(\pmb{\tau})$ denote the expected total reward for the trajectory (sample path) $\pmb{\tau}$ starting from $s_1$, i.e., 
$ R(\pmb{\tau})\equiv\sum^{H-1}_{t=1}\gamma^{t-1}r_t(s_t,a_t)
$, where $\gamma$ is the discount factor and the reward $r_t(s_t,a_t)$ occurring in the $t$-th time step depends on the state $s_t$ and action $a_t$.
Therefore, given 
the process model specified by $\pmb\omega^c$, we are interested in finding the optimal policy maximizing the expected total reward,
\begin{equation}
 \pmb{\pi}_{\pmb{\theta}}^\star(\cdot|\pmb\omega^c) =\arg \max_{\pmb{\pi}_{\pmb{\theta}}}\mu^c(\pmb{\pi}_{\pmb{\theta}}) 
 \equiv 
 \arg \max_{\pmb{\pi}_{\pmb{\theta}}} 
 \E_{\pmb \tau\sim D_{P_{\pmb\omega^c}}^{\pmb{\pi}_{\pmb{\theta}}}(\pmb{\tau})}\left[\left.\sum^{H-1}_{t=1}\gamma^{t-1}r_t\right|\pmb{\pi}_{\pmb{\theta}},s_1\right]. \label{eq: objective1}
\end{equation}


\subsection{Model Risk Quantification and Bayesian Reinforcement Learning}
\label{subsec:ModelRisk}

However, the underlying process model is typically unknown and estimated by the limited historical real-world data.
Here, we focus on Bayesian reinforcement learning (RL) with model risk quantified by the posterior distribution. We consider the \textit{growing-batch RL} setting~\citep{laroche2019multi-batch}. \textit{The process consists in successive periods: In each $p$-th period, a batch of data is collected with a fixed policy from distributed complex bioprocess, it is used to update the knowledge of bioprocess state transition model, and then the policy will be updated for the next period.} 
At any $p$-th period, given all real-world historical data collected so far, denoted by $\mathcal{D}_p$,  
we construct the posterior distribution of state transition model quantifying the model risk, $p(\pmb\omega|\mathcal{D}_p)\propto p(\mathcal{D}_p|\pmb\omega)p(\pmb\omega)$, where the prior $p(\pmb\omega)$ quantifies the existing knowledge on bioprocess dynamic mechanisms.
Since the posterior of previous time period can be the prior for the next update, the posterior will be updated as new process data are collected. There are various \textit{advantages of using the posterior distribution quantifying the model risk}, including: (1) it can incorporate the existing domain knowledge on bioprocess dynamic mechanisms; (2) it is valid even when the historical process data are very limited, which often happens in the biomanufacturing; and (3) it facilitates online learning and  bioprocess knowledge automatic update.

\begin{sloppypar}
At any $p$-th period, to provide the reliable guidance on the dynamic decision making, we need to consider both process inherent stochastic uncertainty and model risk. 
Let $\mu(\pmb{\pi}_{\pmb{\theta}})$ denote the total expected reward accounting for both sources of uncertainty: $\mu(\pmb{\pi}_{\pmb{\theta}}) \equiv \E_{\pmb\omega \sim p(\pmb\omega|\mathcal{D}_p)}\left[\E_{\pmb \tau\sim D_{P_{\pmb\omega}}^{\pmb{\pi}_{\pmb{\theta}}}(\pmb{\tau})}\left[\sum^{H-1}_{t=1}\gamma^{t-1}r_t|\pmb{\pi}_{\pmb{\theta}},s_1, \pmb\omega\right]\right]$, with the inner conditional expectation, $\widetilde{\mu}(\pmb{\pi}_{\pmb{\theta}};\pmb\omega) \equiv\E_{\pmb \tau\sim D_{P_{\pmb\omega}}^{\pmb{\pi}_{\pmb{\theta}}}(\pmb{\tau})}\left[\sum^{H-1}_{t=1}\gamma^{t-1}r_t|\pmb{\pi}_{\pmb{\theta}},s_1, \pmb\omega\right]$ accounting for stochastic uncertainty and the outer expectation accounting for model risk. 
Therefore, given the partial information of bioprocess characterized by $p(\pmb\omega|\mathcal{D}_p)$, we are interested in finding the optimal policy, 
\begin{equation}
 \pmb{\pi}_{\pmb{\theta}}^\star\left(\cdot \left|p(\pmb\omega|\mathcal{D}_p)\right.\right) =\arg \max_{\pmb{\pi}_{\pmb{\theta}}}\mu(\pmb{\pi}_{\pmb{\theta}}) 
 \equiv 
 \arg \max_{\pmb{\pi}_{\pmb{\theta}}} 
 \E_{\pmb\omega \sim p(\pmb\omega|\mathcal{D}_p)}\left[\E_{\pmb \tau\sim D_{P_{\pmb\omega}}^{\pmb{\pi}_{\pmb{\theta}}}(\pmb{\tau})}\left[\left.\sum^{H-1}_{t=1}\gamma^{t-1}r_t
 \right|\pmb{\pi}_{\pmb{\theta}},s_1, \pmb\omega\right]\right].\label{eq: objective2}
\end{equation}
\end{sloppypar}


\section{GREEN SIMULATION ASSISTED REINFORCEMENT LEARNING WITH MODEL RISK} 
\label{sec: green simulation assisted RL with model risk}

In this section, we present the green simulation assisted Bayesian reinforcement learning, which can efficiently leverage the information from historical process trajectory data and accelerate the search for the optimal policy. In Section~\ref{subsec:GreenSimulation_PolicyGradient}, at each $p$-th period and given real-world data $\mathcal{D}_p$, we derive the policy gradient solving the stochastic optimization problem~(\ref{eq: objective2}) and develop the likelihood ratio based green simulation to improve the gradient estimation. Motivated by the metamodel study in~\cite{Dong2018}, 
a decision process mixture proposal distribution and the likelihood ratio based metamodel for policy gradient are derived, which can reuse the process trajectories generated from previous experiments to improve the gradient estimation stability and speed up the search for the optimal policy. In Section~\ref{subsec:algorithms}, we provide the algorithm for proposed online green simulation assisted policy gradient with model risk.  


\subsection{Green Simulation Assisted Policy Gradient}
\label{subsec:GreenSimulation_PolicyGradient}

At each $p$-th period and given real-world data $\mathcal{D}_p$, we develop the green simulation based likelihood ratio to efficiently use the existing process data and facilitate the policy gradient search. Conditional on the posterior distribution $p(\pmb\omega|{\mathcal{D}_p})$, the objective of reinforcement learning is to maximize the expected performance
$\pmb{\pi}_{\pmb{\theta}}^\star\left(\cdot \left|p(\pmb\omega|{\mathcal{D}_p})\right.\right) =\arg \max_{\pmb{\pi}_{\pmb{\theta}}}\mu(\pmb{\pi}_{\pmb{\theta}}).$ Based on eq.~\eqref{eq: objective2}, we can rewrite the objective function,
\begin{eqnarray}
\lefteqn{ \mu(\pmb{\pi}_{\pmb{\theta}})=\E_{\pmb\omega \sim p(\pmb\omega|{\mathcal{D}_p})}\left[\E_{\pmb \tau\sim D_{P_{\pmb\omega}}^{\pmb{\pi}_{\pmb{\theta}}}(\pmb{\tau})}\left[\left.\sum^{H-1}_{t=1}\gamma^{t-1}r_t
 \right|\pmb{\pi}_{\pmb{\theta}},s_1, \pmb\omega\right]\right] }
 \nonumber\\
&=& \int\int\frac{p_{\pmb\omega}(s_1)\prod^{H-1}_{t=1}\pi_{\pmb\theta}(a_t|s_t)p_{\pmb\omega}(s_{t+1}|s_t,a_t)}{p_{\bar{\pmb\omega}}(s_1)\prod^{H-1}_{t=1}\pi_{\bar{\pmb{\theta}}}(a_t|s_t)p_{\bar{\pmb\omega}}(s_{t+1}|s_t,a_t)}p_{\bar{\pmb\omega}}(s_1) \prod^{H-1}_{t=1}\pi_{\bar{\pmb{\theta}}}(a_t|s_t)p_{\bar{\pmb\omega}}(s_{t+1}|s_t,a_t)\sum^{H-1}_{t=1}\gamma^{t-1}r_t p(\pmb\omega|{\mathcal{D}_p}) d\pmb\tau d\pmb\omega 
\nonumber\\
&=& \E_{\pmb\omega \sim p(\pmb\omega|{\mathcal{D}_p})}\left[\E_{\pmb \tau\sim D_{P_{\bar{\pmb\omega}}}^{\pmb{\pi}_{\bar{\pmb{\theta}}}}(\pmb{\tau})}\left[\frac{p_{\pmb\omega}(s_1)\prod^{H-1}_{t=1}\pi_{\pmb\theta}(a_t|s_t)p_{\pmb\omega}(s_{t+1}|s_t,a_t)}{p_{\bar{\pmb\omega}}(s_1)\prod^{H-1}_{t=1}\pi_{\bar{\pmb{\theta}}}(a_t|s_t)p_{\bar{\pmb\omega}}(s_{t+1}|s_t,a_t)}\sum^{H-1}_{t=1}\gamma^{t-1}r_t|\pmb{\pi}_{\pmb{\theta}},s_1,\pmb\omega\right]\right]. \nonumber\\
&=& \E_{\pmb\omega \sim p(\pmb\omega|{\mathcal{D}_p})}\left[\E_{\pmb \tau\sim D_{P_{\bar{\pmb\omega}}}^{\pmb{\pi}_{\bar{\pmb{\theta}}}}(\pmb{\tau})}\left[
\frac{D_{P_{\pmb\omega}}^{\pmb{\pi}_{\pmb{\theta}}}(\pmb{\tau})}{D_{P_{\bar{\pmb\omega}}}^{\pmb{\pi}_{\bar{\pmb{\theta}}}}(\pmb{\tau})}\sum^{H-1}_{t=1}\gamma^{t-1}r_t|\pmb{\pi}_{\pmb{\theta}},s_1,\pmb\omega
\right]\right]. 
\label{eq: total reward with likelihood ratio}
\end{eqnarray}

\noindent The likelihood ratio ${D_{P_{\pmb\omega}}^{\pmb{\pi}_{\pmb{\theta}}}(\pmb{\tau})}/{D_{P_{\bar{\pmb\omega}}}^{\pmb{\pi}_{\bar{\pmb{\theta}}}}(\pmb{\tau})}$
in 
eq. \eqref{eq: total reward with likelihood ratio} can adjust the existing trajectories generated by policy $\pmb{\pi}_{\bar{\pmb{\theta}}}$ and transition model $p(s_{t+1}|s_t,a_t; \bar{\pmb\omega})$ to predict the mean response at the new policy $\mu(\pmb{\pi}_{\pmb{\theta}})$.

Let $k$ denote the \textit{accumulated} number of iterations for the optimal search occurring in the previous $p$ periods. For notation simplification, suppose there is a fixed number of iterations in each period (say $K$). 
At $k$-th iteration, we only generate one posterior sample $\pmb\omega_k \sim p(\pmb\omega|{\mathcal{D}_p})$ to estimate the outer expectation in eq.~(\ref{eq: total reward with likelihood ratio}). 
For the candidate policy $\pmb{\pi}_{\pmb{\theta}_k}$, the likelihood ratio based green simulation is used to estimate the mean response  $\mu(\pmb{\pi}_{\pmb{\theta}_k})$. It can reuse the process trajectories obtained from previous simulation experiments generated by using the policies and state transition models $(\pmb{\pi}_{\pmb{\theta}_i}, \pmb\omega_i)$ with $i=1,2,\ldots,k$. They are obtained in previous $p$ periods with different posterior distributions, i.e., $p(\pmb\omega|{\mathcal{D}_\ell})$ with $\ell=1,2,\ldots,p$. Then, since each proposal distribution is based on a single decision process distribution $D_{P_{{\pmb\omega}_i}}^{\pmb{\pi}_{\pmb{\theta}_i}}(\pmb{\tau})$ specified by $(\pmb{\pi}_{\pmb{\theta}_i}, \pmb\omega_i)$, we create the green simulation \textit{individual likelihood ratio (ILR) estimator} of $\mu(\pmb{\pi}_{\pmb{\theta}_k})$,

\begin{equation}
 \widehat{\mu}^{ILR}_{k,\mathbf{n}}
 \equiv \frac{1}{k}\sum^{k}_{i=1}\frac{1}{n_i}\sum^{n_i}_{j=1}\left[\frac{D_{P_{\pmb\omega_k}}^{\pmb{\pi}_{\pmb{\theta}_k}}(\pmb{\tau}^{(i,j)})}{D_{P_{{\pmb\omega}_i}}^{\pmb{\pi}_{\pmb{\theta}_i}}(\pmb{\tau}^{(i,j)})}\sum^{H_{ij}-1}_{t=1}\gamma^{t-1}r_t(a_t^{(i,j)},s_t^{(i,j)})\right], 
 \mbox{ $\pmb \tau^{(i,j)}\overset{\text{i.i.d}}{\sim} D_{P_{\pmb\omega_i}}^{\pmb{\pi}_{\pmb{\theta}_i}} $}\label{ILR estimator} \nonumber
\end{equation}
where $\pmb \tau^{(i,j)}$ is the $j$-th sample path generated by using $(\pmb{\pi}_{\pmb{\theta}_i}, \pmb\omega_i)$ and $\mathbf{n}=(n_1,n_2,\ldots,n_k)$ is the combination of replications allocated at each $(\pmb{\pi}_{\pmb{\theta}_i}, \pmb\omega_i)$ for $i=1,2,\ldots,k$. Since the process trajectory length is scenario-dependent, we replace the horizon $H$ with $H_{ij}$ to indicate its trajectory dependence.

This expected total reward estimator $ \widehat{\mu}^{ILR}_{k,\mathbf{n}}$ can be used in the policy gradient to search for the optimal policy.
Under some
regularity conditions, we provide the derivation for the policy gradient estimator. 
\begin{eqnarray}
 \lefteqn{ \nabla_{\pmb{\theta}} 
\widetilde{\mu}(\pmb{\pi}_{\pmb{\theta}};\pmb\omega) = \nabla_{\pmb{\theta}}\E_{\pmb \tau\sim D_{P_{\pmb\omega}}^{\pmb{\pi}_{\pmb{\theta}}}}\left[\sum^{H-1}_{t=1}\gamma^{t-1}r_t|\pmb{\pi}_{\pmb{\theta}},s_1,\pmb\omega\right] 
=\int \nabla_{\pmb{\theta}}D_{P_{\pmb\omega}}^{\pmb{\pi}_{\pmb{\theta}}}(\pmb \tau)\left[\sum^{H-1}_{t=1}\gamma^{t-1}r_t(s_t,a_t)\right] d\pmb\tau }
\nonumber\\
&=&\int D_{P_{\pmb\omega}}^{\pmb{\pi}_{\pmb{\theta}}}
(\pmb \tau)
\nabla_{\pmb{\theta}}\mbox{log}(D_{{P_{\pmb\omega}}}^{\pmb{\pi}_{\pmb{\theta}}}(\pmb \tau))\left[\sum^{H-1}_{t=1}\gamma^{t-1}r_t(s_t,a_t)\right] d\pmb\tau \nonumber\\
&=&\int D_{P_{\pmb\omega}}^{\pmb{\pi}_{\pmb{\theta}}}(\pmb{\tau})
\sum^{H-1}_{t=1}\left[\nabla_{\pmb{\theta}}\mbox{log}(\pi_{\pmb\theta}(a_t|s_t)) + \nabla_{\pmb{\theta}}\mbox{log}(p(s_{t+1}|s_t,a_t))\right]\left[\sum^{H-1}_{t=1}\gamma^{t-1}r_t(s_t,a_t)\right] d\pmb\tau \nonumber\\
&=&\int D_{P_{\pmb\omega}}^{\pmb{\pi}_{\pmb{\theta}}}(\pmb{\tau})
\sum^{H-1}_{t=1}\left[\nabla_{\pmb{\theta}}\mbox{log}(\pi_{\pmb\theta}(a_t|s_t))\right]\left[\sum^{H-1}_{t=1}\gamma^{t-1}r_t(s_t,a_t)\right] d\pmb\tau 
\nonumber\\
 &=& \E_{\pmb \tau\sim D_{P_{\pmb\omega}}^{\pmb{\pi}_{\pmb{\theta}}}}\left[
 \left. 
 \sum^{H-1}_{t=1}\nabla_{\pmb{\theta}}\mbox{log}(\pi_{\pmb{\theta}}(a_t|s_t))\left[\sum^{t-1}_{t^\prime=1}\gamma^{{t^\prime}-1}r_{t^\prime}(s_{{t^\prime}},a_{{t^\prime}}) +\sum^{H-1}_{t^\prime=t}\gamma^{{t^\prime}-1}r_{{t^\prime}}(s_{{t^\prime}},a_{{t^\prime}}) \right] 
 \right|
 \pmb{\pi}_{\pmb{\theta}},s_1,\pmb\omega
 \right] 
 \nonumber\\
 &=& 
 \sum^{H-1}_{t=1}\E_{\pmb \tau_{[1:t-1]}}
 \left[ 
 \left.
 \E_{\pmb \tau_{[t:H-1]}}
 \left[
 \left.
 \nabla_{\pmb{\theta}}\mbox{log}(\pi_{\pmb{\theta}}(a_t|s_t))\sum^{t-1}_{t^\prime=1}\gamma^{t^\prime-1}r_{t^\prime}(s_{t^\prime},a_{{t^\prime}}) 
\right|
\pmb{\tau}_{[1:t-1]}\right]
\right|\pmb{\pi}_{\pmb{\theta}},s_1,\pmb\omega
 \right] \nonumber\\
 & & \ \ \ \ + \E_{\pmb \tau\sim D_{P_{\pmb\omega}}^{\pmb{\pi}_{\pmb{\theta}}}}\left[\left.
 \sum^{H-1}_{t=1}\nabla_{\pmb{\theta}}\mbox{log}(\pi_{\pmb{\theta}}(a_t|s_t))\sum^{H-1}_{t^\prime=t}\gamma^{t^\prime-1}r_{t^\prime}(s_{{t^\prime}},a_{{t^\prime}})
 \right|\pmb{\pi}_{\pmb{\theta}},s_1,\pmb\omega \right] \nonumber\\
 &=&\E_{\pmb \tau\sim D_{P_{\pmb\omega}}^{\pmb{\pi}_{\pmb{\theta}}}}\left[ \left. 
 \sum^{H-1}_{t=1}\nabla_{\pmb{\theta}}\mbox{log}(\pi_{\pmb{\theta}}(a_t|s_t))\sum^{H-1}_{t^\prime=t}\gamma^{t^\prime-1}r_{t^\prime}(s_{{t^\prime}},a_{{t^\prime}}) \right|\pmb{\pi}_{\pmb{\theta}},s_1,\pmb\omega\right]\label{eq: future doesn't impact past}\\
 &= &\E_{\pmb \tau\sim D_{P_{\bar{\pmb\omega}}}^{\pmb{\pi}_{\bar{\pmb{\theta}}}}}\left[
 \left. \frac{D_{P_{\pmb\omega}}^{\pmb{\pi}_{\pmb{\theta}}}(\pmb{\tau})}{D_{P_{\bar{\pmb\omega}}}^{\pmb{\pi}_{\bar{\pmb{\theta}}}}(\pmb{\tau})}\sum^{H-1}_{t=1}\nabla_{\pmb{\theta}}\mbox{log}(\pi_{\pmb{\theta}}(a_t|s_t))\sum^{H-1}_{t^\prime=t}\gamma^{t^\prime-1}r_t^\prime({s_{t^\prime},a_{t^\prime}}) \right|\pmb{\pi}_{\pmb{\theta}},s_1,\pmb\omega \right] \label{expected policy graident}
\end{eqnarray}
\begin{sloppypar}

\noindent where eq.~\eqref{expected policy graident} holds due to similar derivation as eq.~\eqref{eq: total reward with likelihood ratio}.
Eq.~\eqref{eq: future doesn't impact past} holds because 

\begin{eqnarray}
\lefteqn{\E_{\pmb \tau_{[1:t-1]}} 
\left. \left[
\left.
\E_{\pmb \tau_{[t:H-1]}}\left[\nabla_{\pmb{\theta}}\mbox{log}(\pi_{\pmb{\theta}}(a_t|s_t))\sum^{t-1}_{t^\prime=1}\gamma^{t^\prime-1}r_{t^\prime}(s_{{t^\prime}},a_{{t^\prime}}) 
\right| {\pmb{\tau}_{[1:t-1]}}\right]
\right|\pmb{\pi}_{\pmb{\theta}},s_1,\pmb\omega\right]}\nonumber \\
&=& \E_{\pmb \tau_{[1:t-1]}} \left.\left[\sum^{t-1}_{t^\prime=1}\gamma^{t^\prime-1}r_{t^\prime}(s_{{t^\prime}},a_{{t^\prime}}) \E_{\pmb \tau_{[t:H-1]}}\left[\nabla_{\pmb{\theta}}\mbox{log}(\pi_{\pmb{\theta}}(a_t|s_t))|{\pmb{\tau}_{[1:t-1]}}\right]
\right|\pmb{\pi}_{\pmb{\theta}},s_1,\pmb\omega\right]
\end{eqnarray}
where
{\begin{eqnarray}
  \lefteqn{\E_{\pmb \tau_{[t:H-1]}}\left[\nabla_{\pmb{\theta}}\mbox{log}(\pi_{\pmb{\theta}}(a_t|s_t))|{\pmb{\tau}_{[1:t-1]}}\right]
     }\nonumber\\
    &=&\prod^{H-1}_{t^\prime=t+1}\int \pi_{\pmb{\theta}}(a_{t^\prime}|s_{t^\prime})p(s_{t^\prime+1}|s_{t^\prime},a_{t^\prime})da_{t^\prime}ds_{t^\prime+1}\int\pi_{\pmb{\theta}}(a_{t}|s_{t})p(s_{t+1}|s_{t},a_{t})\nabla_{\pmb{\theta}}\mbox{log}(p(s_{t+1}|s_{t},a_{t})\pi_{\pmb{\theta}}(a_t|s_t)) da_t ds_{t+1} \nonumber\\
    &=&\int p(s_{t+1},a_t|s_{t})\nabla_{\theta}\mbox{log} p(s_{t+1},a_t|s_{t}) da_t ds_{t+1} \text{, since $p(s_{t+1},a_t|s_{t})=\pi_{\pmb{\theta}}(a_{t}|s_{t})p(s_{t+1}|s_{t},a_{t})$ }\nonumber\\
    &=&\nabla_{\pmb{\theta}}\int p(s_{t+1},a_t|s_{t}) da_t ds_{t+1}=\nabla_{\pmb{\theta}} 1=0.\nonumber
\end{eqnarray}}

By plugging in eq.(\ref{expected policy graident}), the policy gradient 
becomes,
\begin{eqnarray}
 \nabla_{\pmb{\theta}}\mu(\pmb{\pi}_{\pmb{\theta}})&=&\nabla_{\pmb{\theta}}\E_{\pmb\omega}\left[\E_{\pmb \tau\sim D_{P_{\pmb\omega}}^{\pmb{\pi}_{\pmb{\theta}}}(\pmb{\tau})}
 \left[ \left.
 \sum^{H-1}_{t=1}\gamma^{t-1}r_t
\right|\pmb{\pi}_{\pmb{\theta}},s_1, \pmb\omega
 \right]\right]=
\nabla_{\pmb{\theta}}\E_{\pmb\omega}[\widetilde{\mu}(\pmb{\pi}_{\pmb{\theta}};\pmb\omega)] = \E_{\pmb\omega}[\nabla_{\pmb{\theta}}\widetilde{\mu}(\pmb{\pi}_{\pmb{\theta}};\pmb\omega)]
\nonumber \\
 &=&
\E_{\pmb\omega}\left[\E_{\pmb \tau\sim D_{P_{\bar{\pmb\omega}}}^{\pmb{\pi}_{{\bar{\pmb{\theta}}}}}}\left[\sum^{H-1}_{t=1}\nabla_{\pmb{\theta}}\mbox{log}(\pi_{\pmb{\theta}}(a_t|s_t))\frac{D_{P_{\pmb\omega}}^{\pmb{\pi}_{\pmb{\theta}}}(\pmb{\tau})}{D_{P_{\bar{\pmb\omega}}}^{\pmb{\pi}_{\bar{\pmb{\theta}}}}(\pmb{\tau})}\sum^{H-1}_{t^\prime=t}\gamma^{t^\prime-1}r_t^\prime({s_{t^\prime},a_{t^\prime}}) \right]\right]. \label{eq: LR-based policy gradient}
\end{eqnarray}
Then, we obtain the \textit{individual likelihood ratio based policy gradient estimator},
\begin{equation}
 \widehat{\nabla_{\pmb{\theta}}\mu}^{ILR}_{k,\mathbf{n}}
 = \frac{1}{k}\sum^{k}_{i=1}\frac{1}{n_i}\sum^{n_i}_{j=1}\left[\sum^{H-1}_{t=1}\nabla_{\pmb{\theta}}\mbox{log}(\pi_{\pmb{\theta}_k}({a_{t}^{(i,j)}|s_{t}^{(i,j)}}))\frac{D_{P_{{\pmb\omega}_k}}^{\pmb{\pi}_{\pmb{\theta_k}}}(\pmb{\tau}^{(i,j)})}{D_{P_{\pmb\omega_i}}^{\pmb{\pi}_{\pmb{\theta}_i}}(\pmb{\tau}^{(i,j)})}\sum^{H-1}_{t^\prime=t}\gamma^{t^\prime-1}r_t^\prime(a_{t^\prime}^{(i,j)},s_{t^\prime}^{(i,j)})\right].
 \label{eq: LR-based policy gradient estimator}
\end{equation}
\end{sloppypar}

The importance weight or likelihood ratio
${D_{P_{\pmb\omega_k}}^{\pmb{\pi}_{\pmb{\theta}_k}}(\pmb{\tau})}
/{D_{P_{{\pmb\omega_i}}}^{\pmb{\pi}_{{\pmb{\theta}_i}}}(\pmb{\tau})}$ 
is larger for the trajectories $\pmb{\tau}$ that are more likely
to be generated by the policy $\pmb{\pi}_{\pmb{\theta}_k}$ and transition probabilities $P_{\pmb\omega_k}$. 
During the model learning process, the current policy candidate $\pmb{\pi}_{\pmb{\theta}_k}$ can be quite different from the policy $\pmb{\pi}_{{\pmb{\theta}}_i}$ for $i=1,2,\ldots,k-1$ that generated the existing trajectories. Although this importance weight is unbiased, its variance could grow exponentially as the horizon $H$ increases, which restricts their applications.

\textit{Since the likelihood ratio with single proposal distribution can lead to high estimation variance, inspired by the BLR-M metamodel proposed in \cite{Dong2018}, we develop the bioprocess Mixture proposal distribution and Likelihood Ratio based policy gradient estimation (MLR), which allows us to selectively reuse 
the previous experiment trajectories and reduce the gradient estimation variance.} 
Specifically, at the $k$-th iteration of search for optimal policy,  
we generate a posterior sample of process model parameters, $\pmb{\omega}_k\sim p(\pmb\omega|\mathcal{D}_p)$. During the optimal policy search, if there are new process data coming, the posterior will automatically update. The policy candidate $\pmb\pi_{\pmb\theta_k}$ and transition probability model $P(s_{t+1}|s_t,a_t;\pmb\omega_k)$ uniquely define the trajectory distribution $ D_{P_{\pmb\omega_k}}^{\pmb{\pi}_{\pmb{\theta}_k}}(\pmb{\tau})$. 
\textit{Based on the historical trajectories generated during the previous $p$ periods, we create a mixture proposal distribution $\sum_{i=1}^{k} \alpha_i^{k} D_{P_{\pmb\omega_i}}^{\pmb{\pi}_{\pmb{\theta}_i}}(\pmb{\tau})$, and then use it to construct the likelihood ratio,}
\begin{equation}
 f_k(\pmb\tau|\bar{\pmb\theta},\bar{\pmb\omega}) \equiv \frac{D_{P_{\pmb\omega_k}}^{\pmb{\pi}_{\pmb{\theta}_k}}(\pmb{\tau})}
 {\sum_{i=1}^{k} \alpha_i^{k} D_{P_{\pmb\omega_i}}^{\pmb{\pi}_{\pmb{\theta}_i}}(\pmb{\tau})}\label{eq: BLR-M weights 1}
\end{equation}
where $\bar{\pmb\theta}=(\pmb\theta_1,\ldots,\pmb\theta_{k})$, $\bar{\pmb\omega}=(\pmb\omega_1,\ldots,\pmb\omega_{k})$, $\alpha_i^{k}=\frac{n_i}{\sum_{{i}=1}^k n_i}$, and $n_i$ is the number of trajectories generated during the previous $i$-th iteration with $(\pmb{\theta}_i,\pmb{\omega}_i)$ for $i=1,\ldots,k$. 
By replacing the likelihood ratio ${D_{P_{\pmb\omega_k}}^{\pmb{\pi}_{\pmb{\theta}_k}}(\pmb{\tau})}/{D_{P_{{\pmb\omega}_i}}^{\pmb{\pi}_{{\pmb{\theta}_i}}}(\pmb{\tau})}$ in eq.~\eqref{eq: LR-based policy gradient estimator} with $f_k(\pmb\tau|\bar{\pmb\theta},\bar{\pmb\omega})$, the 
green simulation based policy gradient estimator becomes,
\begin{equation}
 \widehat{\nabla_{\pmb{\theta}}\mu}^{MLR}_{k,\mathbf{n}}
 =\frac{1}{k}\sum^{k}_{i=1}\frac{1}{n_i}\sum^{n_i}_{j=1}\left[\sum^{H_{ij}-1}_{t=1}\nabla_{\pmb{\theta}}\mbox{log}(\pi_{\pmb{\theta}_k}(a_t^{(i,j)}|s_t^{(i,j)}))f_{k}(\pmb\tau^{(i,j)}|\bar{\pmb\theta},\bar{\pmb\omega})\sum^{H_{ij}-1}_{t^\prime =t}\gamma^{t^\prime-1}r_{t^\prime}(a_{t^\prime}^{(i,j)},s_{t^\prime}^{(i,j)}) \right] 
 \label{BLR-M sequential gradient estimator}
\end{equation}
where $\pmb \tau^{(i,j)}\overset{\text{i.i.d}}{\sim} D_{P_{\pmb\omega_{i}}}^{\pmb{\pi}_{\pmb{\theta}_{i}}}(\pmb{\tau})$ with $j=1,2,\ldots,n_i$ represent the trajectories generated in the previous $i$-th iteration.
 \textit{Notice that the mixture proposal distribution based likelihoood ratio $f_k(\pmb\tau|\bar{\pmb\theta},\bar{\pmb\omega})$ is bounded by $1/\alpha_i^{k}$. 
 In this way, the mixture likelihood ratio puts higher weight on the existing trajectories that are more likely to be generated by $ D_{P_{\pmb\omega_k}}^{\pmb{\pi}_{\pmb{\theta}_k}}(\pmb{\tau})$ in the $k$-th iteration without assigning extremely large weights on the others.}

Since the parameterization plays an important role in the optimal policy gradient approach, we briefly discuss several possible policy functions. The policy function in reinforcement learning can be either stochastic or deterministic; see~\cite{Silver2014} and \cite{Sutton2018}.
The policy for discrete actions could be defined as the softmax function,
$
 \pi_{\pmb\theta}(a|s) = \frac{e^{\pmb\theta^T\phi(s,a)}}{\sum_{a^\prime\in \mathcal{A}} e^{\pmb{\theta}^T\phi(s,a^{{\prime}})}},
$
where $\phi(s,a) \in\mathbb{R}^d$ is feature vector of state-action pair $(s,a)$. The gradient of the policy function is
$
 \nabla_{\pmb{\theta}}\mbox{log}(\pi_{\pmb\theta}(a|s)) = \phi(s,a) - \sum_{a^\prime \in \mathcal{A}}\phi(s,a^\prime)\pi_{\pmb\theta}(s,a^\prime)
$. For continuous action spaces, we can apply Gaussian policy; say for example 
$
 \pi_{\pmb\theta}(a|s) = \mathcal{N}(\pmb\theta^T\phi(s), \sigma^2)
$
for some constant $\sigma$, where $\phi(s)$ is feature representation of $s$. The gradient of the policy function is
$
 \nabla_{\pmb{\theta}}\mbox{log}(\pi_{\pmb\theta}(a|s)) = \nabla_{\pmb{\theta}} \frac{-(a-\pmb\theta^T\phi(s))^2}{2\sigma^2} = \frac{\pmb\theta^T\phi(s)-a}{\sigma^2}\phi(s)
$. In general, as long as the predictive models have a gradient descent learning algorithm, they can be applied in our approach, such as deep neural network, generalized linear regression, SVM, etc. 
In the empirical study, we considered a two-layer MLP model as our policy function.


 \subsection{Optimal Policy Search Algorithm}
 \label{subsec:algorithms}

 Algorithm~\ref{algo} provides the procedure for the green simulation assisted policy gradient approach to support online learning and guide dynamic decision making.

 \begin{algorithm}[H]
\SetAlgoLined
Input: the number of periods $P$ for real-world dynamic data collection; the number of iterations $K$ for optimal policy search in each period; 
differentiable policy $\pi_{\pmb\theta}(a|s)$, $\forall a \in \mathcal{A}, s \in \mathcal{S}, \pmb\theta \in\mathbb{R}^d$; and initial real-world data $\mathcal{D}_1$. Initialize the set of sample trajectories $\mathcal{E}_{1}$, the set of transition model parameters $\pmb{\Omega}_{1}$, and the set of policy parameters $\pmb{\Theta}_{1}$ to be empty set. 
\\
\For{$p=1,2,\ldots,P$ (at each new real-world data collection point)}{
 \For{$k=(p-1)K+1,(p-1)K+2,\ldots,pK$}{
 1. Generate posterior samples $\pmb{\omega}_{k} \sim p(\pmb{\omega}|\mathcal{D}_p)$ and build the transition model with new parameter $\pmb{\omega}_{k}$, i.e., $p(s_{t+1}|s_t,a_t,\pmb{\omega}_{k})$ for $t=1,2,\ldots,H-1$ \;

 2. Generate $n_k$ trajectories by using the current policy $\pi_{\pmb\theta_k}$ and model parameter $\pmb{\omega}_{k}$\;

 \For{$j=1,2,\ldots,n_k$}{
 (a) Generate $j$-th episode $\pmb{\tau}^{(k,j)}=(s_1^{(k,j)},a_1^{(k,j)},s_2^{(k,j)},a_2^{(k,j)},\ldots, s_{H-1}^{(k,j)},a_{H-1}^{(k,j)},s_H^{(k,j)})$ of state-action sequence starting from initial state $s_1^{(k,j)}\sim p(s_{1}| \pmb{\omega}_{k})$, interacting with transition model
 $s_{t+1}^{(k,j)}\sim p(s_{t+1}|s_t^{(k,j)},a_t^{(k,j)}; \pmb{\omega}_{k})$ and following policy $a_t^{(k,j)}\sim\pi_{\pmb\theta_k}(a_t|s_t^{(k,j)})$ for stochastic policy or $a_t^{(k,j)}=\pi_{\pmb\theta_k}(s_t^{(k,j)})$ for deterministic policy\;
 }\
 3. Reuse the trajectories generated in current and all previous iterations to improve the gradient estimation\;
 \For{$i=1,2,\ldots,k$ {and $j=1,2,\ldots,n_k$}}{
  Construct the mixture proposal distribution based likelihood ratio,
 $f_{k}(\pmb\tau^{(i,j)}|\bar{\pmb\theta},\bar{\pmb\omega})$, by using eq.~\eqref{eq: BLR-M weights 1}.
 }\ 
 4. Calculate the gradient $\widehat{\nabla_{\pmb{\theta}}\mu}^{MLR}_{k,\mathbf{n}}$ based on eq.~\eqref{BLR-M sequential gradient estimator} and update the policy: $\pmb{\theta}_{k+1} \leftarrow \pmb{\theta}_k+ \eta_k\cdot\widehat{\nabla_{\pmb{}}\mu}^{MLR}_{k,\mathbf{n}}$\;
 5. Record new generated trajectories $\mathcal{E}_{k+1}= \mathcal{E}_{k}\cup\{\pmb{\tau}^{(k,j)}| j=1,2,\ldots,n_k\}$, transition model parameters $\pmb{\Omega}_{k+1}=\pmb{\Omega}_{k}\cup\{\pmb{\omega}_{k}\}$ and policy parameters $\pmb{\Theta}_{k+1}=\pmb{\Theta}_{k}\cup\{\pmb{\theta}_k\}$;
 }\
6. Collect new process real-world data $\mathcal{L}_p$ by following the estimated optimal policy $\widehat{\pi}_{\pmb{\theta}_k}^\star(a|{s})$ from Step~(4). Then, update the historical data set $\mathcal{D}_{p+1} = \mathcal{D}_p \cup\mathcal{L}_p$ and the posterior distribution $p(\pmb{\omega}|\mathcal{D}_{p+1})$.
}
\caption{Online Green Simulation Assisted Policy Gradient Policy with Model Risk}\label{algo}
\end{algorithm}

 At any $p$-th period, given the real-world data $\mathcal{D}_{p}$ collected so far, the model risk is quantified by the posterior distribution $p(\pmb{\omega}|\mathcal{D}_{p})$, and then we apply the green simulation assisted policy gradient to search for the optimal policy in Steps~(1)--(4). 
 Specifically, in each $k$-th iteration, we first generate the posterior sample for state transition probability model in Step~(1), $\pmb{\omega}_{k} \sim p(\pmb{\omega}|\mathcal{D}_p)$, and then generate $n_k$ trajectories by using the current policy $\pi_{\pmb\theta_k}$ and model parameter $\pmb{\omega}_{k}$ in Step~(2).  Then, in Steps~(3) and (4), we reuse all historical trajectories and apply the green simulation-based policy gradient to speed up the search for the optimal policy.
 After that, as new real-world data coming, we update the posterior of transition model in Step~(6), and then repeat the above procedure.
In the empirical study, we 
use a fixed learning rate $\eta_k=0.01$. 
\textit{Notice that the proposed mixture likelihood ratio based policy gradient can be easily extended to broader reinforcement learning settings, such as online, offline, and model-free cases.}
 

 
\section{EMPIRICAL STUDY} 
\label{Sec: empirical study}

In this section, we study the performance of MLR using a biomanufacturing example.
The upstream simulation model was built based on a first-principle model proposed by~\cite{Jahic2002} and the downstream chromatography purification process follows~\cite{Martagan2018}. 
The empirical study results show that MLR outperforms the state-of-the-art policy search and baseline model-based stochastic gradient algorithms without BLR-M metamodel.

\subsection{A Biomanufacturing Example}

In this paper, we consider the batch-based biomanfucturing and use the stochastic simulation model built based on our previous study~\citep{wang2019stochastic} to characterize the dynamic evolution of biomanufacturing process. A reinforcement learning model with continuous state and discrete action space is then constructed 
to search for the optimal decisions on chromatography pooling window, which was studied by~\cite{Martagan2018}. Instead of assuming that each chromatography step removes the uniformly distributed random proportion of protein and impurity~\citep{Martagan2018}, we let the random removal fraction following Beta distribution with more realistic and flexible shape. 

This biomanufacturing process consists of: (1) upstream fermentation where cells produce the target protein; and (2) downstream purification to remove the impurities through multiple chromatography
steps. The primary output of fermentation is a mixture including the target protein and significant amount of unwanted impurity
derived from the host cells or fermentation medium.
After fermentation, each batch needs to be purified using
chromatography to meet the specified quality requirements, i.e., purity concentration reaching to certain threshold level $p_d$. 
Since the chromatography typically contributes the main cost for downstream purification, in this paper, we focus
on optimizing the integrated protein purification decisions related to
chromatography operations or pooling window selection. 
To guide the downstream purification dynamic decision making, we formulate the reinforcement learning for biomanufacturing process as follows.

\noindent\textbf{Decision Epoch:} Following~\cite{Martagan2018}, we consider three-step chromatography. During chromatography we observe measurements and make decisions at each decision epoch $\mathcal{T} = \{t: 1, 2, 3\}$. 
 
\noindent\textbf{State Space:} The state $\mathbf{s}_t$ at any decision time $t$ is denoted by the protein-impurity-step tuple $\mathbf{s}_t \triangleq (p_t,i_t,t)$ on the finite space $\mathbf{P} \times \mathbf{I}\times \mathcal{T}$, where $\mathbf{P} \equiv [0, \bar{P}]$ and $\mathbf{I} \equiv [0, \bar{I}]$. 
 The state space $\mathbf{P}$ is bounded by a predefined constant threshold $\bar{P}$ due to limitation in cell viability, growth rate and antibody production rate, etc. 
 The state space $\mathbf{I}$ is bounded by a predefined constant threshold $\bar{I}$ following FDA process quality standards. 

\noindent\textbf{Action Space:} Let $a_t$ denote the selection of pooling window given the state $\pmb{s}_t=(p_t,i_t,t)$ at the time $t\in \mathcal{T}$ following a policy $\pi_{\pmb\theta}(\pmb{s}_t)$. To simplify the problem, we consider 10 candidate pooling windows per chromatography step here.
 
\noindent\textbf{Reward:} At the end of downstream process, we record the reward,

 \begin{equation} \label{eq: rewards}
 r(p_{t},i_{t},t=3) = \begin{cases}
 -c_f, & \text{if $r_t < r_d$},\\
 r(p_d), & \text{if $r_t \geq r_d, p_t \geq p_d$}, \\
 r(p_t) - c_l(p_d-p_t), & \text{if $r_t \geq r_d, p_t \leq p_d$}.
 \end{cases}\nonumber
\end{equation}
We set the failure cost $c_f=\$48$, the protein shortage cost $c_l=\$6$ per milligram (mg), the product price \$5 per mg, $r(p_t)=\$5\times p_t$, the amount of purity percentage requirement $p_d=8$ mg, and the purity requirement $r_d \geq 85\%$. 
The operational cost for each chromatography column is \$8 
for $t\in{1,2,3}$ and $r(p_{t},i_{t},t)=-\$8$ for $t \in\{1,2\}$. 

\noindent\textbf{Initial State:} The random protein and impurity inputs for downstream chromatography are generated with the cell culture first-principle model, which is based on the differential equations with random noise. Here, we consider a fed batch bioreactor dynamic model proposed by~\cite{Jahic2002},
\begin{equation}
 \frac{dX}{dt}=(-\frac{F}{V}+\mu)X \mbox{, }
 \frac{dS}{dt}=\frac{F}{V}(S_i-S)-q_s X \mbox{, }
 P =\nu_1 X 
 \mbox{ and } I=\nu_2 X 
 \label{eq.PDE}
\end{equation}
where $\nu_1\sim \mathcal{N}(0.11, 0.01^2)$ and $\nu_2\sim \mathcal{N}(0.11, 0.01^2)$ denote the constant specific mAb protein production and impurity rates, $X$ denotes the biomass concentration from dry weight $(gL^{-1})$, $V=1000$ is medium volume $(L)$, $S_i\sim \mathcal{N}(780, 40)$ denotes inlet substrate concentration $(gL^{-1})$, $S$ is substrate concentration $(gL^{-1})$, $q_{s,max}=0.57$ is specific maximum rate of substrate consumption $(g\, g^{-1}h^{-1})$, $q_s = q_{s,max}\frac{S}{S+0.1}$ is the specific rate of substrate consumption $(g\,g^{-1}h^{-1})$, $\mu=(q_s-q_m)\cdot Y_{em}$ is the specific growth rate $(h^{-1})$ and $Y_{em}=0.3$ is biomass yield coefficient exclusive maintenance and $q_m=0.013$ is maintenance coefficient $(g\,g^{-1}h^{-1})$. The initial biomass and substrate is set to be ($0\,gL^{-1},40 gL^{-1}$). We set the total time of production fermentation to be 50 days, and obtain $p^{(u)}$ mg of target protein and $i^{(u)}$ mg of impurity by applying the PDEs in (\ref{eq.PDE}). 
After the harvest, we further add the noise, following the normal distribution $\mathcal{N}(0,5^2)$, to account for the overall impact from other factors introduced during the cell production process. Then, the protein $p_1$ and impurity $i_1$ inputs for downstream purification become $p_1\sim\mathcal{N}(p^{(u)},5^2)$ and $i_1\sim\mathcal{N}(i^{(u)},5^2)$. 
Therefore, in the empirical study, this PDE first-principle model based simulation is used to generate the random initial state or input $\mathbf{s}_1=(p_1,i_1,1)$ for downstream chromatography purification.

\noindent\textbf{State Transitions:} In each step of chromatography, the random proportions of protein and impurity will be removed, which depend on the selection of pooling window $a_t$. In specific, given a pooling window, each chromatography step removes random proportions of protein and impurity,
\begin{equation}
 i_{t+1}=(\Psi_t|a_t)i_t \mbox{ and } p_{t+1}=(H_t|a_t)p_t, \nonumber
\end{equation}
where the fraction $\Psi_t|a_t \sim \mbox{Beta}(\psi^l_t|a_t,\psi^u_t|a_t)$ and $H_t|a_t \sim \mbox{Beta}(\eta^l_t|a_t,\eta^u_t|a_t)$ for all $a_t\in \mathcal{A}$ and $t \in \mathcal{T}$.
We use the posterior distribution for model parameters $\psi^l_t|a_t$,$\psi^u_t|a_t$ $\eta^l_t|a_t$ and $\eta^u_t|a_t$ to quantify the model risk. Here we use a uniform prior $\mbox{Unif}(0,300)$ for all parameters and generate the posterior samples based on MCMC using ``PyMC3''.

\noindent\textbf{Policy:} We use a 2-layer perceptron (MLP) of $D=16$ dimensional first layer and 10 dimensional output layer with softmax activation function to parameterize our policy; see Section 11 in~\cite{hastie01statisticallearning} for more discussion. 
For 10 pooling window outputs, there are 10 units $T_{\ell}$ with $\ell=1,\ldots,10$ at the second stage, with the $\ell$-th unit modeling the probability of selecting action $a_{\ell}$ with $\ell=1,\ldots,10$. There are $10$ pooling window candidate actions $a_{\ell}$, $\ell=1,2,\ldots,10$, each being coded as 0-1 variable. The derived feature $Z_d$ depends on the linear combination of the input states $\mathbf{s}$, and then the output $T_{\ell}$ is modeled as a function of linear combinations of the $Z_d$,
\begin{eqnarray}
\small
 Z_d&=&\mbox{Sigmoid}(w_{0d}+\mathbf{w}_d^T \mathbf{s}), d=1,\ldots,D,\nonumber\\
 T_{\ell}&=&\beta_{0\ell}+\pmb\beta_{\ell}^T \mathbf{Z}, \ell=1,\ldots,10\nonumber\\
 \mbox{Prob}(a_{\ell}|\mathbf{s})&\equiv&\mbox{MLP}_{\ell}(\mathbf{s})=g_{\ell}(\mathbf{T}),\ell=1,\ldots,10 
\end{eqnarray}
 where $\mathbf{Z}=(Z_1,Z_2,\ldots,Z_D)$, $\mathbf{T}=(T_1,\ldots,T_{10})$, $\mathbf{w}=(w_{0d},\mathbf{w}_d^T), \pmb{\beta}=(\beta_{0\ell},\pmb\beta_{\ell}^T)$. We can obtain the policy parameters $\pmb\theta=(\mathbf{w},\pmb{\beta})$. The activation function is set to be sigmoid function, i.e., $\mbox{Sigmoid}(x)=\frac{1}{1+e^{-x}}.$ The output function $g_{\ell}(T)$ allows a final transformation of the vector of outputs $\mathbf{T}$, which is set to be softmax functioon $g_{{\ell}}(\mathbf{T})=\frac{e^{T_{\ell}}}{\sum_{\ell=1}^{10}e^{T_{\ell}}}$.

\subsection{Study the Performance of Green Simulation Assisted Policy Gradient}

In this section, we compare the performance of proposed green simulation assisted 
policy gradient with RL (MLR), individual likelihood ratio based policy gradient (ILR) and classical policy gradient (PG).


\begin{itemize}
\item Likelihood ratio based policy gradient with mixture proposal distribution (MLR):
To reduce the computation complexity, instead of reusing all previous iterations, we introduce a rolling window parameter $k_r$ to control how many historical trajectories we use, 
\begin{equation}
\widehat{\nabla_{\pmb{\theta}}\mu}^{MLR}_{k,\mathbf{n}}=\frac{1}{k_r}\sum^{k}_{i=k-k_r+1}\frac{1}{n_i}\sum^{n_i}_{j=1}\left[\sum^{H_{ij}-1}_{t=1}\nabla_{\pmb{\theta}}\mbox{log}(\pi_{\pmb{\theta}_k}(a_t^{(i,j)}|\mathbf{s}_t^{(i,j)}))f_{k}(\pmb\tau^{(i,j)}|\bar{\pmb\theta},\bar{\pmb\omega})\sum^{H_{ij}-1}_{t^\prime =t}\gamma^{t^\prime-1}r_{t^\prime}(a_{t^\prime}^{(i,j)},\mathbf{s}_{t^\prime}^{(i,j)}) \right].
 \label{BLR-M sequential gradient estimator rolling window} \nonumber
\end{equation}

In the empirical study, we use the most recent $k_r=10$ iterations.

\item Likelihood ratio based policy gradient with true transition model known (TLR),
\begin{eqnarray}
 \lefteqn{\widehat{\nabla_{\pmb{\theta}}\mu}^{TLR}_{k,\mathbf{n}}
 =\frac{1}{k_r}\sum^{k}_{i=k-k_r+1}\frac{1}{n_i}\sum^{n_i}_{j=1}\left[\sum^{H_{ij}-1}_{t=1}\nabla_{\pmb{\theta}}\mbox{log}(\pi_{\pmb{\theta}_k}(a_t^{(i,j)}|\mathbf{s}_t^{(i,j)}))f_{k}(\pmb\tau^{(i,j)}|\bar{\pmb\theta},\pmb\omega^c)\sum^{H_{ij}-1}_{t^\prime =t}\gamma^{t^\prime-1}r_{t^\prime}(a_{t^\prime}^{(i,j)},\mathbf{s}_{t^\prime}^{(i,j)}) \right] } \nonumber\\
 &=& 
 \frac{1}{k_r}\sum^{k}_{i=k-k_r+1}\frac{1}{n_i}\sum^{n_i}_{j=1}\left[\sum^{H_{ij}-1}_{t=1}\nabla_{\pmb{\theta}}\mbox{log}(\pi_{\pmb{\theta}_k}(a_t^{(i,j)}|\mathbf{s}_t^{(i,j)})) \frac{\prod^{H-1}_{t=1}\pi_{\pmb\theta_k}(a_t|\mathbf{s}_t)}{\sum_{i=1}^k\prod^{H-1}_{t=1}\pi_{\pmb{\theta}_i}(a_t|\mathbf{s}_t)}\sum^{H_{ij}-1}_{t^\prime =t}\gamma^{t^\prime-1}r_{t^\prime}(a_{t^\prime}^{(i,j)},\mathbf{s}_{t^\prime}^{(i,j)}) \right],
 \nonumber 
\end{eqnarray}

\noindent where 
the last step holds because $\frac{p_{\pmb\omega^c}(s_1)\prod^{H-1}_{t=1}\pi_{\pmb\theta_k}(a_t|\mathbf{s}_t)p_{\pmb\omega^c}(\mathbf{s}_{t+1}|\mathbf{s}_t,a_t)}{\sum_{i=1}^k p_{\pmb\omega^c}(s_1)\prod^{H-1}_{t=1}\pi_{\pmb{\theta}_i}(a_t|\mathbf{s}_t)p_{\pmb\omega^c}(\mathbf{s}_{t+1}|\mathbf{s}_t,a_t)}=\frac{\prod^{H-1}_{t=1}\pi_{\pmb\theta_k}(a_t|\mathbf{s}_t)}{\sum_{i=1}^k\prod^{H-1}_{t=1}\pi_{\pmb{\theta}_i}(a_t|\mathbf{s}_t)}$.

\item Individual likelihood ratio based policy gradient (ILR): It is obtained based on Equation~\eqref{eq: LR-based policy gradient estimator}, 
$$ 
 \widehat{\nabla_{\pmb{\theta}}\mu}_{k,\mathbf{n}}^{ILR}
 = \frac{1}{k}\sum^{k}_{i=1}\frac{1}{n_i}\sum^{n_i}_{j=1}\left[\sum^{H-1}_{t=1}\nabla_{\pmb{\theta}}\mbox{log}(\pi_{\pmb{\theta}_k}(a_t|\mathbf{s}_t))\frac{D_{P_{{\pmb\omega}_k}}^{\pmb{\pi}_{\pmb{\theta_k}}}(\pmb{\tau}^{(i,j)})}{D_{P_{\pmb\omega_i}}^{\pmb{\pi}_{\pmb{\theta}_i}}(\pmb{\tau}^{(i,j)})}\sum^{H-1}_{t^\prime=t}\gamma^{t^\prime-1}r_t^\prime(a_{t^\prime}^{(i,j)},\mathbf{s}_{t^\prime}^{(i,j)})\right].$$
 
 \item Empirical policy gradient (PG): It uses the point estimator of state transition model parameter as the true one,
 \begin{equation}
 \widehat{\nabla_{\pmb{\theta}}\mu}^{PG}= \frac{1}{n_i}\sum^{n_i}_{j=1}\left[\sum^{H-1}_{t=1}\nabla_{\pmb{\theta}}\mbox{log}(\pi_{\pmb{\theta}}(a_t|\mathbf{s}_t))\sum^{H-1}_{t^\prime=t}\gamma^{t^\prime-1}r_t^\prime(a_{t^\prime}^{(i,j)},\mathbf{s}_{t^\prime}^{(i,j)})\right]. \label{eq: policy gradient estimator} \nonumber
\end{equation}
\end{itemize}

Notice that in MLR, ILR, PG approaches, the underlying state transition model is unknown and estimated by finite real-world data. In TLR, we assume the model is known.

\begin{figure}[!h]
 \centering
 \subfloat[$n_i=50$]{{\includegraphics[width=0.5\textwidth]{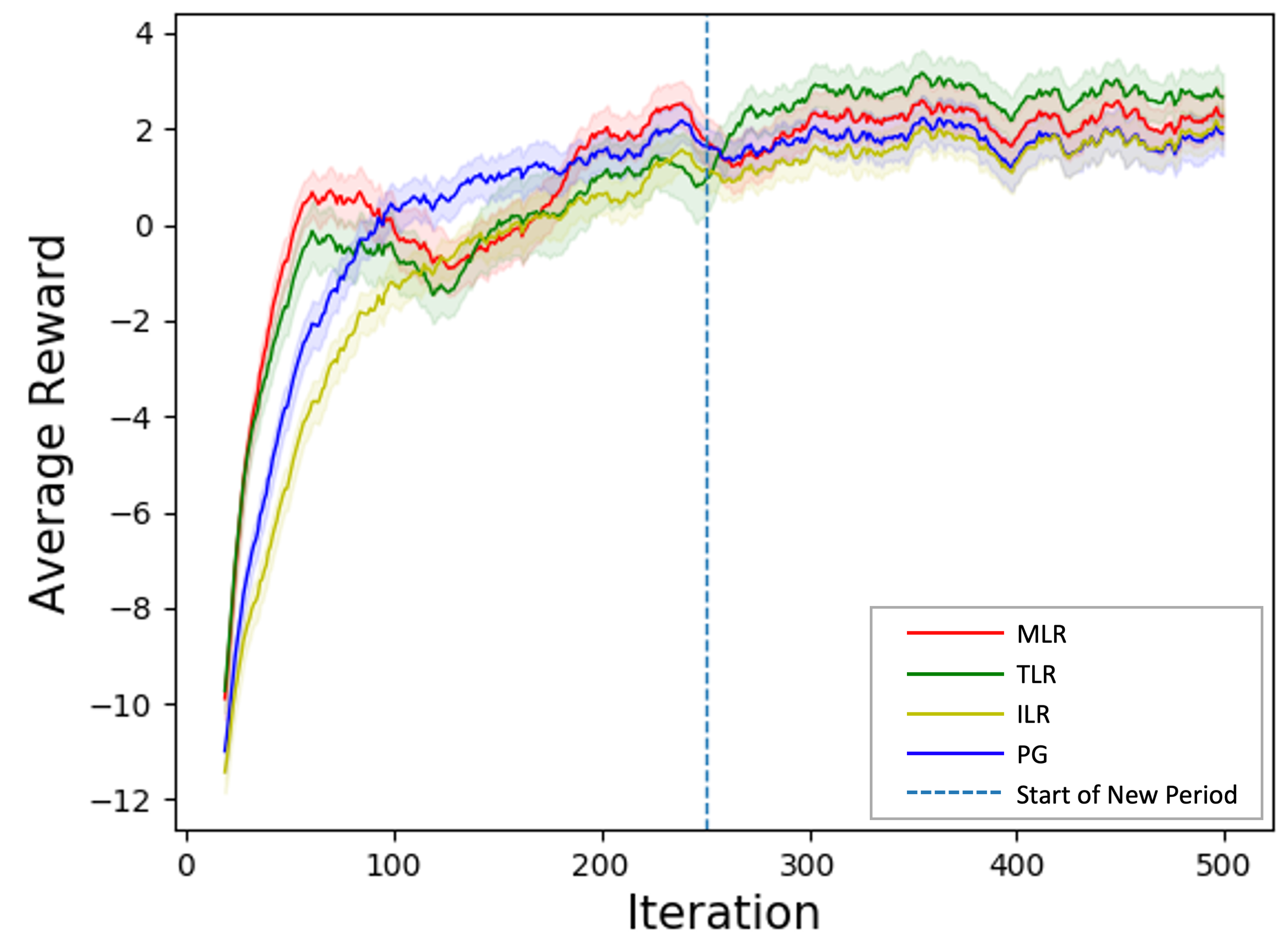} }\label{plot: converagence nr=20 ni=50}}
 \subfloat[$n_i=25$]{{\includegraphics[width=0.5\textwidth]{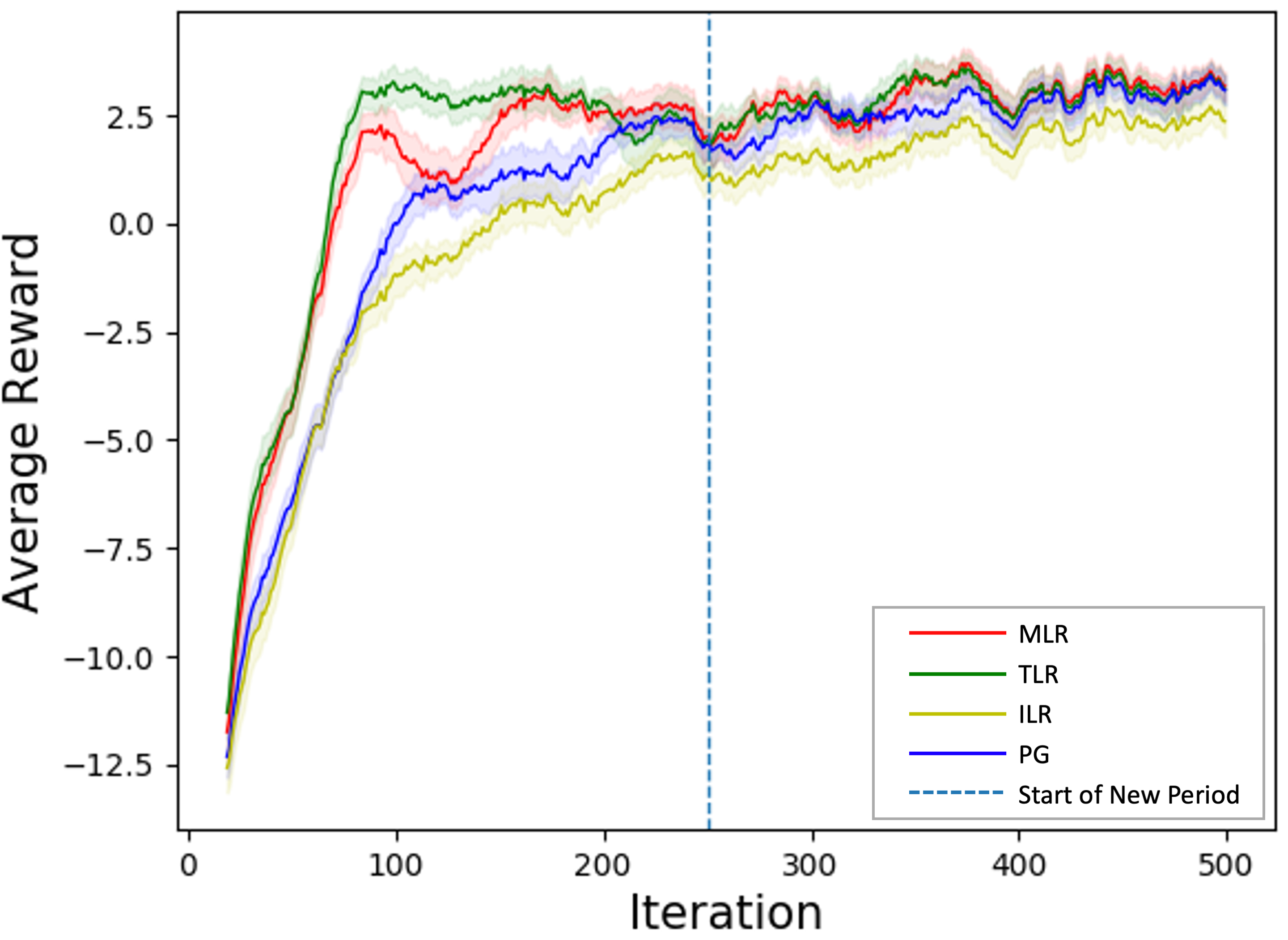} }\label{plot: converagence nr=20 ni=25}}
 \vspace{-1.2em}
 \subfloat[$n_i=10$]{{\includegraphics[width=0.5\textwidth]{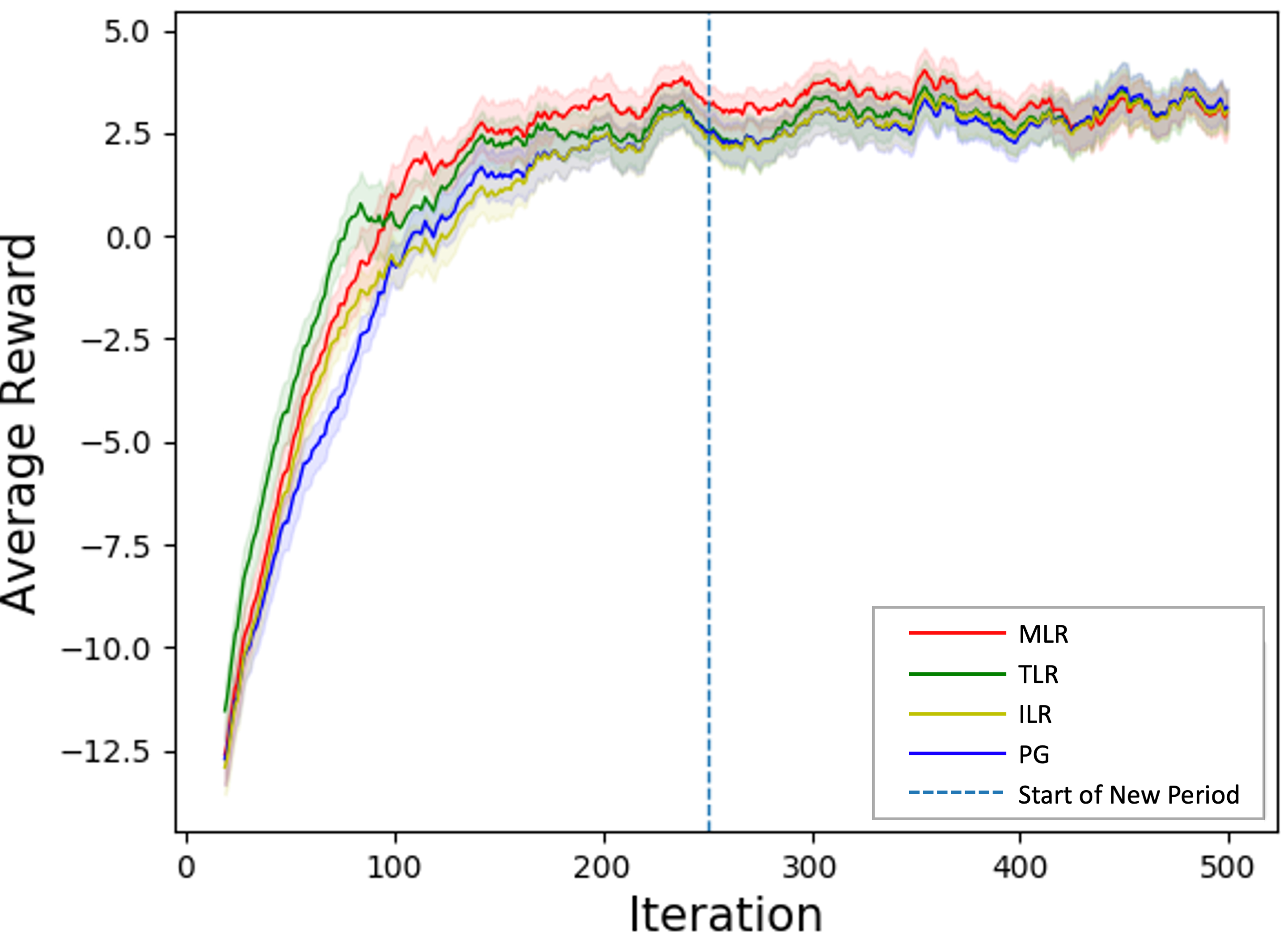} }\label{plot: converagence nr=20 ni=10}} %
 \subfloat[$n_i=5$]{{\includegraphics[width=0.5\textwidth]{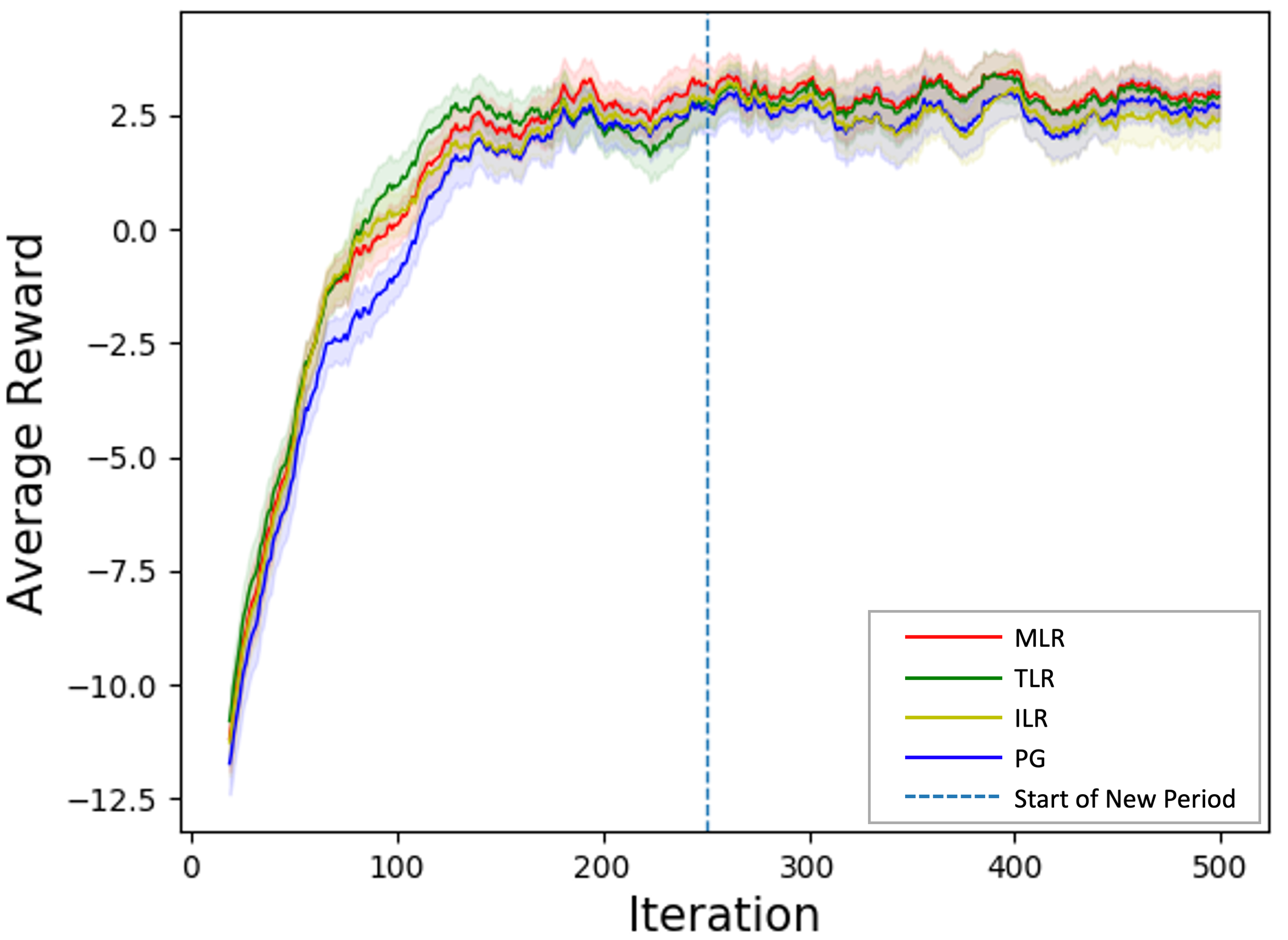} }\label{plot: converagence nr=20 ni=5}}%
 \caption{Convergence results of MLR, TLR, ILR and PG. 
 }%
 \label{plot: converagence}
\end{figure}

Here we set the amount of real-world data $m=20$ for chromatography operation.
Fig.~\ref{plot: converagence} shows the convergence performance of MLR, TLR, ILR, and PG. The results are based on $M=5$ macro replications. 
The x-axis represents the iteration index $k$, and the vertical dash line indicates the time when the new real-world process data are collected. Let ${r}_{h}(k)$ denote the average reward of the policy obtained from the $k$-th iteration in the $h$-th macro replications, which is estimated by running $r_{test}=200$ trajectories with the true state transition model.
The y-axis reports $\bar{r}(k)=\frac{1}{M}\sum_{h=1}^M {r}_{h}(k)$.
We also plot the 95\% confidence band for each approach, $[\bar{r}(k)-1.96\times \mbox{SE}(\bar{r}(k)), \bar{r}(k)+1.96\times \mbox{SE}(\bar{r}(k))]$, where $\mbox{SE}(\bar{r}(k))=\frac{1}{\sqrt{M(M-1)}}\sqrt{\sum^M_{h=1}({r}_h(k)-{\bar{r}}(k))^2}$.
Fig.~\ref{plot: converagence} 
 shows that MLR (red line) converges faster than PG and ILR. 
 To better compare the performance of candidate algorithms, we apply the common random numbers (CRNs) for each macro replication.

From Fig~\ref{plot: converagence}, we can see the algorithms have already converged after 400 iterations. 
We compare the performance of policies obtained from MLR, PG and ILR based on the results from the last 100 iterations. We record the sample mean $\mu_a=\frac{1}{100}\sum_{k=401}^{500}\bar{r}(k)$ and standard error $\mbox{SE}=\frac{1}{10}\sqrt{\frac{1}{99}\sum_{k=401}^{500}(\bar{r}(k) -\mu_a)^2}$ in Table~\ref{table: result}.
The results show that MLR tends to have better performance than both PG and ILR approaches.


\begin{table}[!h]
\small
\centering
\caption{{Average reward estimated based on last 100 iterations for MLR, TLR, ILR and PG.
}\label{table: result}}
\begin{tabular}{l|cc|cc|cc|cc}
\hline
\multirow{2}{*}{} & \multicolumn{2}{c|}{$n_i=50$} & \multicolumn{2}{c|}{$n_i=25$} & \multicolumn{2}{c|}{$n_i=10$} & \multicolumn{2}{c}{$n_i=5$} \\ \cline{2-9} 
 & Mean & SE & Mean & SE & Mean & SE & Mean & SE \\ \hline
MLR & 2.23 & 0.10 & 3.25 & 0.09 & 3.07 & 0.09 & 2.92 & 0.11 \\
TLR & 2.75 & 0.10 & 3.14 & 0.09 & 3.08 & 0.09 & 2.83 & 0.11 \\
PG & 1.80 & 0.09 & 3.04 & 0.10 & 3.10 & 0.10 & 2.53 & 0.11 \\
ILR & 1.83 & 0.10 & 2.36 & 0.10 & 3.01 & 0.10 & 2.39 & 0.13 \\ \hline
\end{tabular}
\end{table}

{When $n_i=25$ based on $M=5$ macro replications, the average runtime for MLR is 53.0 mins (12.6 mins for updating posterior distribution and 40.4 mins for policy search). The average runtime for PG is 34.3 mins (12.9 mins for updating posterior distribution and 21.4 mins for policy search). The average runtime for ILR is 50.1 mins (12.2 mins for updating posterior distribution and 37.9 mins for policy search).}

\section{CONCLUSIONS}\label{Sec: conclusion}

We propose a green simulation assisted policy gradient algorithm. It can reduce the policy gradient estimation variance through selectively reusing the experiment data and automatically allocating more weight to those historical trajectories that are more likely generated by the stochastic decision process of interest. In addition, since we quantify the state transition probabilistic model risk with the posterior distribution, our model-based reinforcement learning can simultanesouly support online learning and guide dynamic decision making. Thus, the proposed approach is robust to model risk, and it can be applicable to various cases with different amounts of real-world data and process dynamic knowledge. In this paper, the empirical study of biomanufacturing example is used to illustrate that our approach can perform better than the state-of-art reinforcement learning and policy gradient approaches. 

\bibliographystyle{plainnat}

\bibliography{wscbib}

\begin{thebibliography}{11}
\providecommand{\natexlab}[1]{#1}
\providecommand{\url}[1]{\texttt{#1}}
\expandafter\ifx\csname urlstyle\endcsname\relax
  \providecommand{\doi}[1]{doi: #1}\else
  \providecommand{\doi}{doi: \begingroup \urlstyle{rm}\Url}\fi

\bibitem[{Dong} et~al.(2018){Dong}, {Feng}, and {Nelson}]{Dong2018}
J.~{Dong}, M.~B. {Feng}, and B.~L. {Nelson}.
\newblock Unbiased metamodeling via likelihood ratios.
\newblock In \emph{2018 Winter Simulation Conference (WSC)}, pages 1778--1789,
  Dec 2018.
\newblock \doi{10.1109/WSC.2018.8632506}.

\bibitem[Feng and Staum(2017)]{FengGreenSim2017}
Mingbin Feng and Jeremy Staum.
\newblock Green simulation: Reusing the output of repeated experiments.
\newblock \emph{ACM Transactions on Modeling and Computer Simulation (TOMACS)},
  27\penalty0 (4):\penalty0 23:1--23:28, October 2017.
\newblock ISSN 1049-3301.
\newblock \doi{10.1145/3129130}.
\newblock URL \url{http://doi.acm.org/10.1145/3129130}.

\bibitem[Hastie et~al.(2001)Hastie, Tibshirani, and
  Friedman]{hastie01statisticallearning}
Trevor Hastie, Robert Tibshirani, and Jerome Friedman.
\newblock \emph{The Elements of Statistical Learning}.
\newblock Springer Series in Statistics. Springer New York Inc., New York, NY,
  USA, 2001.

\bibitem[Jahic et~al.(2002)Jahic, Rotticci-Mulder, Martinelle, Hult, and
  Enfors]{Jahic2002}
M.~Jahic, J.~Rotticci-Mulder, M.~Martinelle, K.~Hult, and S.-O. Enfors.
\newblock Modeling of growth and energy metabolism of pichia pastoris producing
  a fusion protein.
\newblock \emph{Bioprocess and Biosystems Engineering}, 24\penalty0
  (6):\penalty0 385--393, 2002.
\newblock ISSN 1615-7605.
\newblock \doi{10.1007/s00449-001-0274-5}.
\newblock URL \url{https://doi.org/10.1007/s00449-001-0274-5}.

\bibitem[Laroche and Tachet~des Combes(2019)]{laroche2019multi-batch}
Romain Laroche and Remi Tachet~des Combes.
\newblock Multi-batch reinforcement learning.
\newblock In \emph{The 4th Multidisciplinary Conference on Reinforcement
  Learning and Decision Making (RLDM)}, July 2019.
\newblock URL
  \url{https://www.microsoft.com/en-us/research/publication/multi-batch-reinforcement-learning/}.

\bibitem[Martagan et~al.(2018)Martagan, Krishnamurthy, Leland, and
  Maravelias]{Martagan2018}
Tugce Martagan, Ananth Krishnamurthy, Peter~A. Leland, and Christos~T.
  Maravelias.
\newblock Performance guarantees and optimal purification decisions for
  engineered proteins.
\newblock \emph{Operations Research}, 66\penalty0 (1):\penalty0 18–41,
  January 2018.
\newblock ISSN 0030-364X.
\newblock \doi{10.1287/opre.2017.1661}.
\newblock URL \url{https://doi.org/10.1287/opre.2017.1661}.

\bibitem[Mnih et~al.(2015)Mnih, Kavukcuoglu, Silver, Rusu, Veness, Bellemare,
  Graves, Riedmiller, Fidjeland, Ostrovski, Petersen, Beattie, Sadik,
  Antonoglou, King, Kumaran, Wierstra, Legg, and Hassabis]{mnih2015humanlevel}
Volodymyr Mnih, Koray Kavukcuoglu, David Silver, Andrei~A. Rusu, Joel Veness,
  Marc~G. Bellemare, Alex Graves, Martin Riedmiller, Andreas~K. Fidjeland,
  Georg Ostrovski, Stig Petersen, Charles Beattie, Amir Sadik, Ioannis
  Antonoglou, Helen King, Dharshan Kumaran, Daan Wierstra, Shane Legg, and
  Demis Hassabis.
\newblock Human-level control through deep reinforcement learning.
\newblock \emph{Nature}, 518\penalty0 (7540):\penalty0 529--533, February 2015.
\newblock ISSN 00280836.
\newblock URL \url{http://dx.doi.org/10.1038/nature14236}.

\bibitem[Schaul et~al.(2016)Schaul, Quan, Antonoglou, and
  Silver]{Schaul2016PrioritizedER}
Tom Schaul, John Quan, Ioannis Antonoglou, and David Silver.
\newblock Prioritized experience replay.
\newblock \emph{CoRR}, abs/1511.05952, 2016.

\bibitem[Silver et~al.(2014)Silver, Lever, Heess, Degris, Wierstra, and
  Riedmiller]{Silver2014}
David Silver, Guy Lever, Nicolas Heess, Thomas Degris, Daan Wierstra, and
  Martin Riedmiller.
\newblock {Deterministic Policy Gradient Algorithms}.
\newblock In \emph{Proceedings of the 31st International Conference on Machine
  Learning, ICML}, Beijing, China, June 2014.
\newblock URL \url{https://hal.inria.fr/hal-00938992}.

\bibitem[Sutton and Barto(2018)]{Sutton2018}
Richard~S. Sutton and Andrew~G. Barto.
\newblock \emph{Reinforcement Learning: An Introduction}.
\newblock A Bradford Book, Cambridge, MA, USA, 2018.
\newblock ISBN 0262039249.

\bibitem[Wang et~al.(2019)Wang, Xie, Martagan, Akcay, and
  Corlu]{wang2019stochastic}
Bo~Wang, Wei Xie, Tugce Martagan, Alp Akcay, and Canan~G. Corlu.
\newblock Stochastic simulation model development for biopharmaceutical
  production process risk analysis and stability control.
\newblock In \emph{Proceedings of the 2019 Winter Simulation Conference}. IEEE,
  Inc., 2019.

\end{thebibliography}

\section*{AUTHOR BIOGRAPHIES}

\noindent {\bf HUA ZHENG} is Ph.D. student of the Department of Mechanical and Industrial Engineering (MIE) at Northeastern University. His
research interests includes machine learning, data analytics, computer simulation and stochastic optimization. His email address is \href{zheng.hua1@husky.neu.edu}{zheng.hua1@husky.neu.edu}. 

\noindent {\bf WEI XIE} is an assistant professor in MIE at Northeastern University.
She received her M.S. and Ph.D. in Industrial Engineering and Management Sciences (IEMS) at Northwestern University. Her research
interests include interpretable Artificial Intelligence (AI), computer simulation, data analytics,  stochastic optimization, and blockchain development for cyber-physical system risk management, learning, and automation.
Her email address is \href{w.xie@northeastern.edu}{w.xie@northeastern.edu}. Her website is \href{http://www1.coe.neu.edu/~wxie/}{http://www1.coe.neu.edu/$\sim$wxie/}

\noindent {\bf BEN MINGBIN FENG } is an assistant professor in actuarial science at the University of Waterloo. He earned his Ph.D. in IEMS at Northwestern University. 
His research interests include stochastic simulation design and analysis, optimization via simulation, nonlinear optimization, and financial and actuarial applications of simulation and optimization methodologies. His e-mail address is \href{ben.feng@uwaterloo.ca}{ben.feng@uwaterloo.ca}. His website is \href{http://www.math.uwaterloo.ca/~mbfeng/}{http://www.math.uwaterloo.ca/$\sim$mbfeng/}. \\

\end{document}